\documentclass[11pt,a4paper]{article}
\usepackage{colacl, epic, eepic, epsfig}
\title{Measuring efficiency in high-accuracy,\\ broad-coverage statistical parsing\thanks{Thanks to everyone in the Brown Laboratory for Linguistic Information
Processing (BLLIP) for valuable discussion on these issues.  This
research was supported in part by NSF IGERT Grant \#DGE-9870676, and NSF LIS Grant \#SBR-9720368.}}
\author{Brian Roark \And Eugene Charniak
\AND
{\rm Brown Laboratory for Linguistic Information Processing (BLLIP), and}
\AND
{\rm Cognitive and Linguistic Sciences}\\Box 1978\\Brown University\\Providence, RI  02912\\{\tt brian-roark@brown.edu}
\And
{\rm Computer Science}\\Box 1910\\Brown University\\Providence, RI  02912\\{\tt ec@cs.brown.edu}}
\begin{document}
\maketitle
\abstract 
Very little attention has been paid to the comparison of efficiency
between high accuracy statistical parsers.  This paper proposes one
machine-independent metric that is general enough to allow comparisons 
across very different parsing architectures.  This metric, which we
call ``events considered'', measures the number of ``events'', however 
they are defined for a particular parser, for
which a probability must be calculated, in order to find the parse.
It is applicable to single-pass or multi-stage parsers.  We discuss
the advantages of the
metric, and demonstrate its usefulness by using it to compare two
parsers which differ in several fundamental ways.

\section{Introduction}
The past five years have seen enormous improvements in
broad-coverage parsing accuracy, through the use of statistical
techniques.  The parsers that perform at the highest level of
accuracy (Charniak \shortcite{Charniak97,Charniak00}; Collins
\shortcite{Collins97,Collins00}; \egcite{Ratna97} use
probabilistic models with a very large number of parameters, which can 
be costly to use in evaluating structures.  Parsers
that have been  built for this level of accuracy have generally
been compared only with respect to accuracy, not efficiency.  This is
understandable: their great selling point is the high level of
accuracy they are able to achieve.  In addition, these parsers are
difficult to compare with respect to efficiency: the models are
quite diverse, with very different kinds of parameters and different
estimation and smoothing techniques.  Furthermore, the 
search and pruning strategies have been 
quite varied, from beam-search to best-first, and with different
numbers of distinct stages of processing.

At a very general level, however, these approaches share some key
characteristics, and it is at this general level that we would like to 
address the issue of efficiency.  In each of these approaches, scores
or weights are calculated for events, e.g. edges or other structures,
or perhaps constituent/head or even head/head relations. The scores
for these events are compared and 
``bad'' events, i.e. events with relatively low scores, are either 
discarded (as in beam search) or sink to the bottom of the heap
(as in best-first).  In fact, this general characterization is
basically 
what goes on at each of the stages in multi-stage parsers, although
the events that are being weighted, and the models by which they are
scored, may change\footnote{Even within the same stage, events can be
heterogeneous.  See the discussion of the EC parser below.}.  In each
parser's final stage, the parse which emerges with the
best score is returned for evaluation. 

We would like to propose an efficiency metric which we call {\it
events considered\/}.  An event is {\it considered\/} when a score is
calculated for it.  Search and pruning techniques can be judged to 
improve the
efficiency of a parser if they reduce the number of events that must
be considered en route to parses with the same level of accuracy.
Because an event must have a score for a statistical parser to decide
whether it should be retained or discarded, there is no way to
improve this number without having improved either
the efficiency of the search (through, say, dynamic programming) or the 
efficacy of the pruning.  We will argue that this is not the case with 
competitor measures, such as time or total heap operations, which can
be improved through optimization techniques that do not change the
search space.  This is not to say that these techniques do not have a
great deal of value; simply that, for comparisons between approaches
to statistical parsing, the implementations of which may or may not
have carried out the same optimizations, they are less informative
than the metric we have proposed.

Some recent papers on efficiency in statistical parsing have
looked at the number of pops from a heap as the relevant measure of
efficiency \cite{Caraballo98,Charniak98,Blaheta99}, and have
demonstrated techniques for improving the scoring function so that
this number is dramatically reduced.  This is also a score that cannot 
be ``artificially'' reduced through optimization.
It may very well be, however,
that some significant part of a parser's function is not an operation
on a heap.  For example, a parser could run a part-of-speech ({\small POS})
tagger over the string as a first stage.  What is relevant for this
first stage are the number of ({\small POS},word) pairs that must be
considered by the tagger.  Each of these pairs 
would have a score calculated for them, and would hence be an {\it
event considered\/}.  The events in the second stage may be, for
example, edges in the chart.  A parser's efficiency score would be the
total number of these considered events across all stages.

The principle merits of this metric are that it is general enough to
cover different search and pruning techniques (including 
exhaustive parsing); that it is machine-independent; and that it
is, to a certain extent, implementation-independent.  The last of
these might be what recommends the metric most, insofar as it is
not the case for other simple metrics.  For example, using
time as a metric is perfectly general, and there are
ways to normalize for processor differences (see \egcite{Moore00b}.
However, unless one is 
comparing two implementations that are essentially identical in all
incidental ways, it is not possible to normalize for certain specifics 
of the implementation.  For example, how probabilities
are accessed, upon which processing time is very dependent, can differ 
from implementation to implementation (see discussion below).  Thus, while
time may be ideal for highly controlled studies of relatively similar
algorithms (as in \egcite{Moore00a}, its applicability for comparing
diverse parsers is problematic.

Let us consider a specific example: calculating scores from highly
conditioned, interpolated probability distributions.  First we will
discuss conditional probability models, followed by an illustration of 
interpolation.

A simple probabilistic context free grammar ({\small PCFG}) is a
context free grammar with a probability assigned to each rule: the
probability of the righthand side of the rule given the lefthand side
of the rule.  These probabilities can be estimated via their relative
frequency in a corpus of trees.  For instance, we can assign a
probability to the rule {\small S} $\rightarrow$ {\small NP VP} by
counting the number of occurrences of this rule in the corpus, and
dividing by the total number of {\small S} nodes in the corpus.  We
can improve the probability model if we add in more conditioning
events beyond the lefthand side of the rule.  For example, if we throw 
in the parent of the lefthand side in the tree within which it appears, we can immediately see a dramatic
improvement in the maximum likelihood parse \cite{Johnson98b}.  That
is, instead of:

\vspace*{-.2in}
\begin{small}
\begin{eqnarray}
P(RHS|LHS) &=& \frac{P(LHS,RHS)}{P(LHS)}\nonumber
\end{eqnarray}
\end{small}
the probability of the rule instance is:
\begin{small}
\begin{eqnarray}
P(RHS|LHS,P_{LHS}) &=& \frac{P(LHS,RHS,P_{LHS})}{P(LHS,P_{LHS})}\nonumber
\end{eqnarray}
\end{small}
where $P_{LHS}$ is the parent above the lefthand side of the rule.
This additional conditioning event allows us to capture the fact that
the distribution of, say, {\small S} node expansions underneath {\small VPs}
is quite different than that of {\small S} nodes at the root
of the tree.  The models that we will be discussing in this
paper condition on many such events, somewhere between five and ten.
This can lead to sparse data problems, necessitating some kind of
smoothing - in these cases, deleted interpolation.

The idea behind deleted
interpolation \cite{Jelinek80} is simple: mix the empirically observed 
probability using $n$ conditioning events with lower order models.  The 
mixing coefficients, $\lambda_{n}$, are functions of the frequency of
the joint occurrence of the conditioning events, estimated from a held 
out portion of the corpus.  Let $e_{0}$ be
the event whose probability is to be conditioned, $e_{1}\ldots e_{n}$
the $n$ conditioning events used in the model, and $\hat{P}$ the
empirically observed conditional probability.  Then the following is a 
recursive definition of the interpolated probability:

\vspace*{-.2in}
\begin{small}
\begin{eqnarray}
P(e_{0}|e_{1}\ldots e_{n}) &=&
\lambda_{n}(e_{1}\ldots e_{n})\hat{P}(e_{0}|e_{1}\ldots e_{n}) + \hspace*{.2in}\nonumber\\
&&({\mathrm 1-}\lambda_{n}(e_{1}\ldots e_{n}))P(e_{0}|e_{1}\ldots e_{n-1})\nonumber \label{eq:int}
\end{eqnarray}
\end{small}
This has been shown to be very
effective in circumstances where sparse data requires smoothing to
avoid assigning a probability of zero to a large number of possible
events that happen not to have been observed in the training data
with the $n$ conditioning events.

Using such a model\footnote{The same points hold for other smoothing
methods, such as backing off.}, the time to calculate a particular
conditional probability can be 
significant.  There are a variety of techniques that can be used to
speed this up, such as pre-compilation or caching.  These 
techniques can have a fairly large effect on the time of computation,
but they contribute little to a comparison between pruning techniques 
or issues of search.  More generally, optimization and lack of it is something
that can obscure algorithm similarities or differences, over and above
differences in machine or platform.  Researchers whose interest lies
in improving parser accuracy might not care to improve the efficiency once
it reaches an acceptable level.  This should not bar us from trying to 
compare their techniques with regards to efficiency.

Another such example contrasts our metric with
one that measures total heap operations.  Depending on the pruning
method, it might be possible to evaluate an event's probability
and throw it away if it falls below some threshold, rather than
pushing it onto the heap.  Another option in 
the same circumstance is to simply push all analyses
onto the heap, and let the heap ranking decide if they ever surface
again.  Both have their respective time trade-offs (the cost of
thresholding versus heap operations), and which is
chosen is an implementation issue that is orthogonal to the relative
search efficiency that we would like to evaluate.

In contrast to time or total heap operations, there is no incidental optimization 
that allows the parser to avoid calculating scores for analyses.  A
statistical parser that prunes the search space cannot perform this
pruning without scoring events that must be either retained or discarded.  
A reduction in events considered without a loss of accuracy counts as
a novel search or pruning 
technique, and as such should be explicitly evaluated as a competitor
strategy.  The basic point that we are making here is that our metric
measures that which is central to statistical parsing techniques, and
not something that can be incidentally improved.

In the next section, we outline two quite different statistical parsers,
and present some results using our new metric.

\section{Comparing statistical parsers}
To illustrate the utility of this metric for comparing the efficiency of
radically different approaches to broad-coverage parsing, we will
contrast some results from a two-stage best-first parser
\cite{Charniak00} with a single-pass left-to-right, incremental
beam-search parser \cite{Roark00b}.  Both of these parsers (which we
will refer to, henceforth, as the {\small EC} and {\small BR} parsers,
respectively) score between 85 and 90 percent average precision and
recall; both condition the probabilities of events on a large number
of contextual parameters in more-or-less the way outlined above;  and
both use 
boundary statistics to assign partial structures a figure-of-merit,
which is the product of the probability of the structure in its own right
and a score for its likelihood of integrating with its surrounding
context.  

Both of the parsers also use parameterized pruning strategies, which
will be described when the parsers are outlined.  Results will be
presented for each parser at a range of parameter values, to give a
sense of the behavior of the parser as more or fewer events are taken
into consideration.  From this data, we shall be able to see the
degree to which the events considered score correlates with time, as
well as the convergence in accuracy.  

The parsers were trained on sections 2-21 and tested on section 23 of
the Penn Wall St. Journal Treebank \cite{Marcus93}, which are the
standards in the statistical parsing literature.
Accuracy is reported in terms of average labelled
precision and recall.  Precision is the number of correct constituents
divided by the number of constituents proposed by the parser.  Recall
is the number of correct constituents divided by the number of
constituents in the actual parse.  Labelled precision and recall
counts only non-part-of-speech non-terminal constituents.  The two
numbers are generally quite close, and are averaged to give a single
composite score.

\subsection{EC parser}
\begin{table}[t]
\begin{small}
\begin{tabular} {|p{.7in}|p{.55in}|p{.7in}|p{.55in}|}
\hline
\multicolumn{4}{|c|}{section 23: 2416 sentences of length $\leq$
100}\\
\multicolumn{4}{|c|}{Average length: 23.46 words/sentence}\\\hline
{Times past first parse} & {Avg. Prec/Rec} &
{Events Considered${}^{\dag}$} & {Time in seconds${}^{\dag}$}\\\hline
{21} & {89.7} & {212,014} & {26.7}\\\hline
{13} & {89.6} & {107,221} & {14.0}\\\hline
{7.5} & {89.1} & {48,606} & {6.7}\\\hline
{2.5} & {86.8} & { 9,621} & {1.5}\\\hline
{2} & {85.6} & {6,826} & {1.1}\\\hline
\multicolumn{4}{l}{\footnotesize ${}^{\dag}$per sentence}
\end{tabular}
\end{small}
\caption{Results from the {\small EC} parser at different initial
parameter values}\label{tab:res1}
\end{table}

The {\small EC} parser first prunes the search space by building a
chart containing only the most likely edges.  Each new edge is assigned a
figure-of-merit ({\small FOM}) and pushed onto a heap.  The {\small
FOM} is the product of the probability of the constituent given the
simple {\small PCFG} and the boundary statistics.  Edges that are
popped from the heap are put into the chart, and standard chart
building occurs, with new edges being pushed onto the heap.  This
process continues until a complete parse is found; hence this is a 
best-first approach.  Of course, the chart building does not
necessarily need to stop when the first parse is found;  it can
continue until some stopping criterion is met.  The
criterion that was used in the trials that will be reported here is
a multiple of the number of edges that were present in the chart when
the first parse was found.  Thus, if the parameter is 1, the parser
stops when the first parse is found; if the parameter is 10, the
parser stops when the number of edges in the chart is ten times the
number that were in the chart when the first parse was found.  

\begin{table*}[t]
\begin{center}\begin{small}
\begin{tabular} {|p{.8in}|p{.6in}|p{.7in}|p{.7in}|p{.7in}|}
\hline
\multicolumn{5}{|c|}{section 23: 2416 sentences of length $\leq$
100}\\
\multicolumn{5}{|c|}{Average length: 23.46 words/sentence}\\\hline
{Base Beam Factor} & {Avg. Prec/Rec} &
{Events Considered${}^{\dag}$} & {Time in seconds${}^{\dag}$} &
{Pct. failed}\\\hline
{$10^{-12}$} & {85.9} & {265,509} & {7.6} & {1.3} \\\hline
{$10^{-11}$} & {85.7} & {164,127} & { 4.3} & {1.7} \\\hline
{$10^{-10}$} & {85.3} & {100,439} & { 2.7} & {2.2} \\\hline
{$10^{-8}$} & {84.3} & { 36,861} & { 0.9} & {3.8} \\\hline
{$10^{-6}$} & {81.8} & { 13,512} & { 0.4} & {7.1} \\\hline
\multicolumn{4}{l}{\footnotesize ${}^{\dag}$per sentence}
\end{tabular}
\end{small}\end{center}
\caption{Results from the {\small BR} parser at different initial
parameter values}\label{tab:res2}
\end{table*}

This is the first stage of the parser.  The second stage takes all of
the parses packed in the chart that are above a certain probability
threshold given the {\small PCFG}, and assigns a score using the full
probability model.  To evaluate the probability of each parse, the
evaluation proceeds from the top down.  Given a particular constituent, it
first evaluates the probability of the part-of-speech of the head of
that constituent, conditioned on a variety of contextual information
from the context.  Next, it evaluates the probability of the head
itself, given the part-of-speech that was just predicted (plus other
information).  Finally, it evaluates the probability of the rule
expansion, conditioned on, among other things, the {\small POS} of the 
head and the head.  It then moves down the tree to evaluate the newly
predicted constituents.  See \newcite{Charniak00} for more details on
the specifics of the parser.

Notice that the events are heterogeneous.  One of the key events in the 
model is the constituent/head relation, which is not an edge.  Note
also that this two-stage search strategy means that many edges will
be considered multiple times, once by the first stage and in every
complete parse within which they occur in the second stage, and hence
will be counted multiple times by our metric. 

The parse with the best score is returned for evaluation in terms of
precision and 
recall.  Table \ref{tab:res1} shows accuracy and efficiency results
when the {\small EC} parser is run at various initial parameter
values, i.e. the number of times past the first parse the
first-stage of the parser continues.

\subsection{BR parser}
The {\small BR} parser proceeds from left-to-right across the string,
building analyses top-down in a single pass.  While its accuracy is
several points 
below that of the {\small EC} parser, it is useful in circumstances
requiring incremental processing, e.g. on-line speech recognition, 
where a multi-stage parser is not an option.

Very briefly, partial analyses are ranked
by a figure-of-merit that is the product of their probability (using the
full conditional probability model) and a look-ahead probability,
which is a measure of the likelihood of the current stack state of an
analysis rewriting to the look-ahead word at its left-corner.
Partial analyses are popped from the heap, expanded, and pushed back
onto the heap.  When an analysis is found that extends to the
look-ahead word, it is pushed onto a new heap, which collects these
``successful'' analyses until there are ``enough'', at which point the 
look-ahead is moved to the next word in the string, and all of the
``unsuccessful'' analyses are discarded.  This is a
beam-search, and the criterion by which it is judged that ``enough''
analyses have succeeded can be either narrow (i.e. stopping early) or
wide (i.e. stopping late).  The unpruned parse with the highest
probability that successfully covers the entire input string is
evaluated for accuracy.  

The beam parameter in the trials that will be reported here, is
called the base beam factor, and it works as follows.  Let $\beta$ be
the base beam factor, and let $\tilde{p}$ be the probability of the
highest ranked ``successful'' parse.  Then any analysis whose
probability falls below $\alpha\beta\tilde{p}$, where $\alpha$ is the
cube of the number of successful analyses, is discarded.  The basic
idea is that we want the beam to be very wide if there are few
analyses that have extended to the current look-ahead word, but
relatively narrow if many such analyses have been found.  Thus, if
$\beta = 10^{-12}$, and 100 analyses have extended to the current
look-ahead word, then a candidate analysis must have a probability
above $10^{-6}\tilde{p}$ to avoid being pruned.  After 1000
candidates, the beam has narrowed to $10^{-3}\tilde{p}$.  
Table \ref{tab:res2} shows accuracy and efficiency results
when the {\small BR} parser is run at various base beam factors.  
See \newcite{Roark00b} for more details on the specifics of this
parser.

The conditional probability model that is used in the {\small BR}
parser is constrained by the left-to-right nature of the algorithm.
Whereas the conditional probability model used in the second stage of
the {\small EC} parser has access to the full parse trees, and thus
can condition the structures with information from either the left or
right context, any model used in the
{\small BR} parser can only use information from the left-context,
since that is all that has been built at the moment the probability of 
a structure is evaluated.  For example, a subject {\small NP} can be
conditioned on the head of 
the sentence (usually the main verb) in the {\small EC} parser, but
not in the {\small BR} parser, since the head of sentence has yet to
be encountered.  This accounts for some of the accuracy 
difference between the two parsers.  Also, note that the {\small BR}
parser can and does fail to find a parse in some percentage of cases,
as a consequence of the incremental beam-search.  This percentage is
reported as well.

\section{Discussion}
The number of ways in which these two parsers differ is large, and
many of these 
differences make it difficult to compare their relative efficiency.  
A partial list of these complicating differences is the following:
\begin{itemize}
\item Best-first vs. beam search pruning strategy, which impacts the
number of events that must be retained
\item Two-stage vs. single pass parsing
\item Heterogeneous events, within and between parsers
\item Different conditional probability models, with different numbers 
of conditioning events, and slightly different methods of
interpolation
\item {\small EC} parser written in C{\small ++}; {\small BR} parser
written in C
\end{itemize}

In addition, for these runs, the {\small EC} parser parallelized the
processing by sending each sentence individually off to different
processors on the network, whereas the {\small BR} parser was run on a
single computing server. Since for the {\small EC} parser we do not
know which sentence went to 
which processor, nor how fast each individual processor was,  time is a
particularly poor point of comparison.

\begin{figure}
\epsfig{file=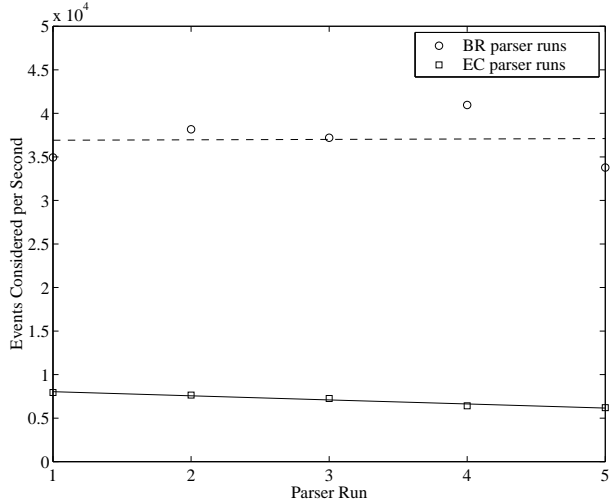, width = 3.2in}
\caption{Events Considered per Second for each parser run, with a
linear fit}\label{fig:vs1}
\end{figure}

\begin{figure}
\epsfig{file=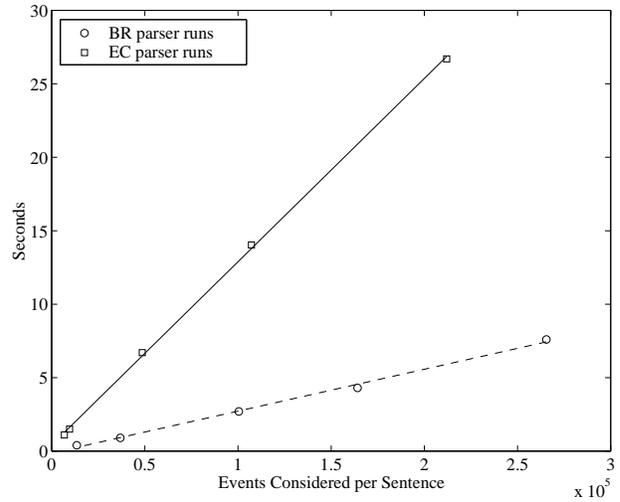, width = 3.2in}
\caption{Time vs. Events Considered per Sentence, with a linear fit}\label{fig:vs2}
\end{figure}

In order for our metric to be useful, however, it should
be highly correlated with time.  Figure \ref{fig:vs1} shows the number 
of events considered divided by the total parse time for each of the
five runs reported for each parser.  While there is some noise between 
each of the runs, this ratio is relatively constant across the runs, as shown by the linear fit,
indicating a very high correlation between the number of events
considered and the total time.  Figure \ref{fig:vs2} plots the edges
considered versus time per sentence for all of the runs reported in the
tables above, and the linear fit for each is drawn as well.  As we can
see from both plots, number of events considered is a good proxy
measure for time in both parsers.  

\begin{figure}
\epsfig{file=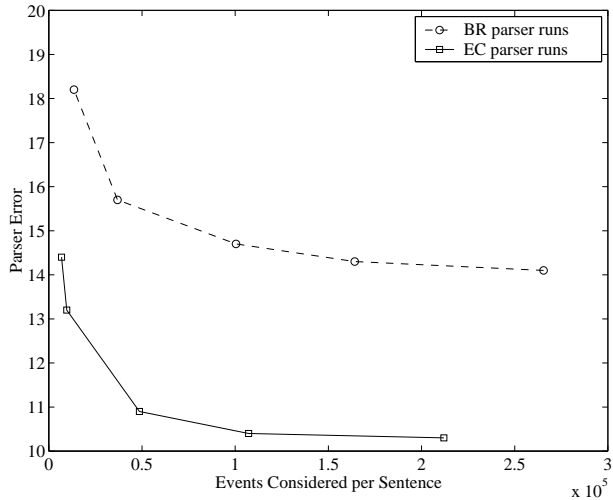, width = 3.2in}
\caption{Reduction in parser error as the number of events considered increases}\label{fig:vs3}
\end{figure}

Now the question is how to judge the relative efficiency using this
measure.  Given that both parsers are parameterized, the number of
events considered can be made essentially arbitrarily high or
arbitrarily low.  We should thus look at the performance of the
parsers over  a range of parameter values.  Figure \ref{fig:vs3} shows the
convergence in accuracy of the models, as more and more events are
considered.  The improvement in accuracy in the graph is represented
as a reduction in parser error, i.e. 100 - average precision/recall.
Both of the parsers show a fairly similar pattern of convergence to
their respective minimum errors.

Given this information, there are two directions that one can go.  The 
first is to simply take this information at face value and make
judgments about the relative efficiency on the basis of these numbers.
We may, however, want to take the comparison one step further, and
look at how quickly each parser converges to its respective best
accuracy, regardless of what that best accuracy is.  In a sense,
this would focus the evaluation on the search aspects of the parser,
apart from the overall quality of the probability model.

\begin{figure}
\epsfig{file=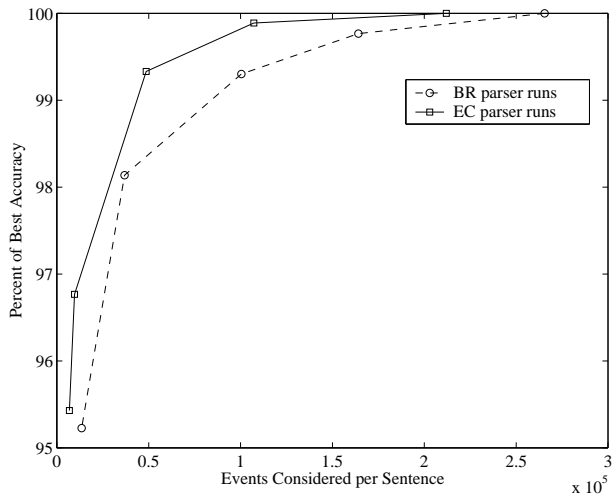, width = 3.2in}
\caption{Convergence to highest accuracy parse as number of events considered increase}\label{fig:vs4}
\end{figure}

Figure \ref{fig:vs4} plots the percentage of the highest
accuracy parse achieved versus the number of events considered.  The
convergence of the {\small BR} parser lies to the right of the
convergence of the {\small EC} parser, indicating that the {\small EC} 
parser takes fewer
edges considered to converge on the best possible accuracy given the
model.  Notice that both parsers had runs with approximately 100,000
events considered, but that the {\small EC} parser is within .1
percent of the best accuracy (basically within noise) at that point, while the
{\small BR} parser still has a fair amount of improvement to go before 
reaching the best accuracy.  Thus the {\small EC} parser needs to
consider fewer events to find the best parse.

This is hardly surprising given what we know about the pruning
strategies.  The first stage of the {\small EC} parser uses dynamic
programming techniques on the chart to evaluate edges only once.  The
{\small BR} parser, in contrast, must evaluate constituents once for
every parse within which they occur.  Particularly useless
constituents will be thrown out once by the {\small EC} parser, but
perhaps many times by the {\small BR} parser.

This difference in efficiency is tangible, but it is relatively
small.  What would be problematic in this domain would be orders of
magnitude differences, which we don't get here.

\section{Conclusion}
We have presented in this paper a very general, machine- and
implementation-independent metric that can be used to compare the
efficiency of quite different statistical parsers.  To illustrate its
usefulness, we compared the performance of two parsers that follow
different strategies in arriving at their parses, and which on the
surface would appear to be very difficult to compare with respect to
efficiency.  
Despite this, the two algorithms seem to require a fairly similar number 
of events considered to squeeze the most accuracy out of their
respective models. 
Furthermore, the decrease in events considered in both cases was
accompanied by a more-or-less proportional decrease in time.  This
data confirmed our intuitions that the two algorithms are roughly
similar in terms of efficiency.  It also lends support to
consideration of this metric as a legitimate, machine and
implementation independent measure of statistical parser efficiency.

In practice, the scores on this measure could be reported alongside of 
the standard {\small PARSEVAL} accuracy measures \cite{Black91}, as an
indicator of the amount of work required to arrive at the parse.  What 
is this likely to mean to researchers in high accuracy, broad-coverage
statistical parsing?  Unlike accuracy measures, whose fluctuations of
a few tenths of percent are attended to with interest, such an
efficiency score is likely to be attended to only if there is an order 
of magnitude difference.  On the other hand, if two parsers have very
similar performance in accuracy, the relative efficiency of one over the other may recommend its use.

When can this metric be used to compare parsers?  We would contend
that it can be used whenever measures such as precision and recall
can be used, i.e. same training and testing corpora.  If the parser is 
working in an entirely different search space, such as with a
dependency grammar, or when the training or testing portions of the
corpus are different, then it is not clear that such comparisons
provide any insight into the relative merits of different parsers.
Much of the statistical parsing literature has settled on specific
standard training and testing corpora, and in this circumstance, this
measure should be useful for evaluation of efficiency.

In conclusion, our efficiency metric has tremendous generality, and is
tied to the operation of statistical parsers in a way that recommends
its use over time or heap operations as a measure of efficiency.  
\bibliographystyle{fullname}
\bibliography{ber}
\end{document}